\documentclass{llncs}
\usepackage{subcaption}
\usepackage{amsmath,amssymb,amsfonts}
\usepackage[compress]{cite}
\usepackage{balance}
\usepackage{algorithmic}
\usepackage{algorithm}
\usepackage{graphicx}
\usepackage{pgfplots}
\usepackage{textcomp}

\begin{document}
\frontmatter          % for the preliminaries
\pagestyle{headings}  % switches on printing of running heads
\addtocmark{LSTM} % additional mark in the TOC
\mainmatter              % start of the contributions

\title{Ensemble of LSTMs and feature selection for human action prediction
}
\titlerunning{LSTM for HAP}  % abbreviated title (for running head)
\author{Tomislav Petkovi\'c, Luka Petrovi\'c, Ivan Markovi\'c \and Ivan Petrovi\'c}
\authorrunning{Tomislav Petković et al.} % abbreviated author list (for running head)
\tocauthor{}
\institute{University of Zagreb Faculty of Electrical Engineering and Computing\\
Laboratory for Autonomous Systems and Mobile Robotics (LAMOR)\\
\email{\{petkovic, luka.petrovic, ivan.markovic, ivan.petrovic\}@fer.hr}
}

\maketitle

\begin{abstract} As robots are becoming more and more ubiquitous in human environments, it will be necessary for robotic systems to better understand and predict human actions.
However, this is not an easy task, at times not even for us humans, but based on a relatively structured set of possible actions, appropriate cues, and the right model, this problem can be computationally tackled.
In this paper, we propose to use an ensemble of long-short term memory (LSTM) networks for human action prediction.
To train and evaluate models, we used the MoGaze\footnote{\url{https://humans-to-robots-motion.github.io/mogaze/}} dataset -- currently the most comprehensive dataset capturing poses of human joints and the human gaze.
We have thoroughly analyzed the MoGaze dataset and selected a reduced set of cues for this task.
Our model can predict (i) which of the labeled objects the human is going to grasp, and (ii) which of the macro locations the human is going to visit (such as table or shelf).
We have exhaustively evaluated the proposed method and compared it to individual cue baselines.
The results suggest that our LSTM model slightly outperforms the gaze baseline in single object picking accuracy, but achieves better accuracy in macro object prediction.
Furthermore, we have also analyzed the prediction accuracy when the gaze is not used, and in this case, the LSTM model considerably outperformed the best single cue baseline.
\end{abstract}
\section{Introduction}
\label{sec:intro}
With the robots becoming more sophisticated, their presence in human environments increases and includes close collaboration with humans.
Such applications involve proximate interaction of robots and humans, yielding novel challenges concerning system efficiency and human safety.
Robots that physically share space with humans will need to understand and learn by interacting and observing humans \cite{kratzer2020mogaze}.
However, this is not an easy task, at times, not even for us humans.
Hence, human action and full-body motion prediction are becoming increasingly important research topics for the robotics research community \cite{rudenko2020human}.
\begin{figure}[t!]
  \centering
\includegraphics[width=\textwidth, height = 6cm]{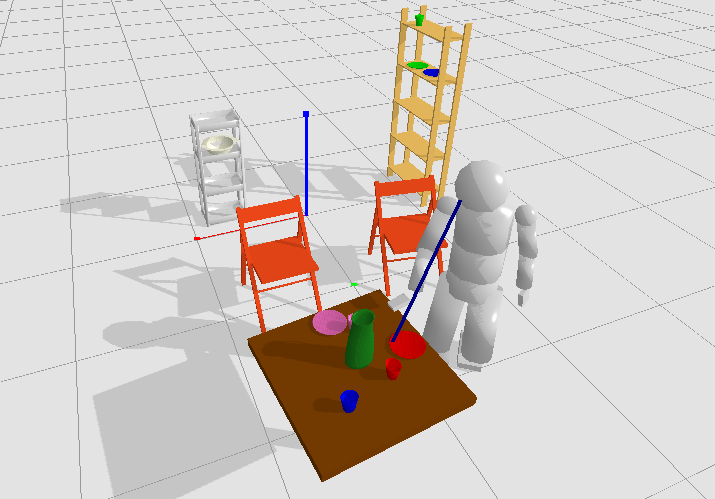}
  \centering
\caption{Visualization of the MoGaze datased based on PyBullet\cite{coumans2016pybullet}. The subject is about to pick the red plate from the table which is strongly indicated by the subject's gaze illustrated with the blue line.}
\label{fig:mogaze}
\vspace{-0.5cm}
\end{figure}

In recent years, human action prediction applications ranged from robotized warehouses \cite{petkovic2019human, petkovic2020human} to sedentary object-picking domain \cite{schydlo2018anticipation, Huang2015, shi2019you} and full-body motions \cite{li2016online, kratzer2020anticipating, zhang2017view}.
Common approaches to solving the human action prediction problem are based on hidden Markov models \cite{Kelley2008} or  conditional random fields \cite{wang2006hidden}, which try to learn observed patterns.
In works \cite{Bader2009, Huang2015, kratzer2020mogaze, shi2019you} authors have indicated that the eye-gaze is a powerful predictor of human action.
They tackle the problem using machine learning models such as support vector machine \cite{Huang2015} or recurrent neural networks \cite{kratzer2020mogaze}.
In \cite{shi2019you} authors calculate the similarity between the hypothetical gaze points on the objects and the actual gaze points and use the nearest neighbor algorithm to classify the intended object.

In the last few years, multiple datasets concerning motion and action prediction have become publicly available but, to the best of our knowledge, none of them couple these two problems.
Examples of purely motion prediction datasets are: KITTI \cite{geiger2012we}, ETH \cite{pellegrini2009you} and UCY \cite{lerner2007crowds}.
We encourage the reader to examine Table 2 in \cite{rudenko2020human} for a detailed listing.
These datasets offer enough diverse data to train and test human motion prediction models focused on answering the question \textit{"Where is the human going to be during the next $N$ steps?"}, but they are not adequately labeled with the context which would help to answer \textit{"What is (the goal of) the observed human motion?}".
On the other hand, datasets tailored for models focused on the second question, like the CMU's motion capture database \cite{CMUDB} and G3D \cite{bloom2012g3d} that excel in action diversity, but they are focused on distinguishing between different actions (jumping, punching, sitting), do not incorporate complicated motion patterns, and usually are not long enough for a long term human motion prediction problem.
The MoGaze \cite{kratzer2020mogaze} dataset  (Fig.~\ref{fig:mogaze}) positions itself as an excellent blend of the aforementioned datasets because all the recorded motion has a labeled purpose (an object picking).
Its subset has already been used by the authors for human motion prediction problems based on RNNs and trajectory optimization \cite{kratzer2020anticipating, kratzer2020prediction}.
Therein, they used Euclidean distance of the right hand to each object as an action prediction signal, improving their original motion prediction result.

In this paper, we rely on findings in \cite{kratzer2020anticipating, kratzer2020mogaze, shi2019you} and propose a human action prediction model based on the ensemble of recurrent neural networks, namely long-short term memory (LSTM) networks.
To train and evaluate our model, we used and thoroughly analyzed the MoGaze dataset and selected a reduced set of features for human action prediction.
Our model can predict (i) which of the labeled objects the human is going to grasp, and (ii) which of the macro locations the human is going to visit (such as table or shelf).
Experimental results are focused on comparative performance with respect to individual cue accuracy, especially the gaze which is a powerful action predictor, and demonstrate the effectiveness of the proposed model.

\section{Proposed Human Action Prediction}
\label{sec:method}
We design the proposed action prediction model as a general model for full-body motion that successfully captures relations between input cues and picked objects, but without learning specific relations between objects in a dataset.
Another important aspect that needs to be captured by an action prediction model are long-term dependencies, since goal inferring cues appear usually much earlier than the actual picking action.
For example, imagine a human who is given the task of picking a specific object.
First, they would look towards the object and start walking to it.
While walking, the gaze of the human would not be fixed only on the object, but would also wander around the scene, especially if there are obstacles to be negotiated.
Given that, a well-designed human action prediction model should take into account that the gaze becomes fixed at the object early in the sequence and can wonder thereafter.
In other words, to successfully infer the goal, the model should be able to remember the most important past cue values, e.g. early gaze fixation at the object, as well as capture local tendencies, such as human approaching the object.
In order to achieve that, in the ensuing section we propose an ensemble of long short-term memory (LSTM) networks that also take distances of human joints towards each of the goals.
However, each joint in the input feature set adds to the complexity and the network parameter number, which not only increases the run-time, but can also impede the training process.
Consequently, we further introduce a feature selection method based on signal correlations and individual effectiveness to act as an action prediction cue.

\subsection{Ensembles of LSTMs for Action Prediction}
\label{sec:Ensembles}
A recurrent neural network (RNN) is a feed forward network derivate that allows previous network outputs to be used as its inputs with the introduction of a hidden state.
By memorizing past states, RNN is able to process sequences of inputs  \cite{dupond2019thorough}.
An LSTM \cite{hochreiter1997lstm} is an RNN extension that takes advantage of its gates, thus enabling it to capture long-term dependencies by partially solving the vanishing gradient problem.
They have been used in a plethora of sequence regression and classification problems, such as trajectory prediction \cite{zhang2019sr}, speech emotion recognition \cite{zhao2019speech} and action prediction \cite{si2019attention}.
As previously stated, one of the fundamental requirements for our model is to avoid learning relations between objects specific to a dataset and to be able to remember the most important past cue values.
For example, in the MoGaze dataset the objects are placed on three macro locations, two shelves and a table, which do not move during the experiments.
If we give the model, e.g. distances to all the goals as an input, the model could implicitly learn relations between those macro locations that would not hold for other datasets.
Furthermore, our model should also be capable of handling a varying number of goals.

\begin{figure}[t!]
  \centering
\includegraphics[width =\textwidth ]{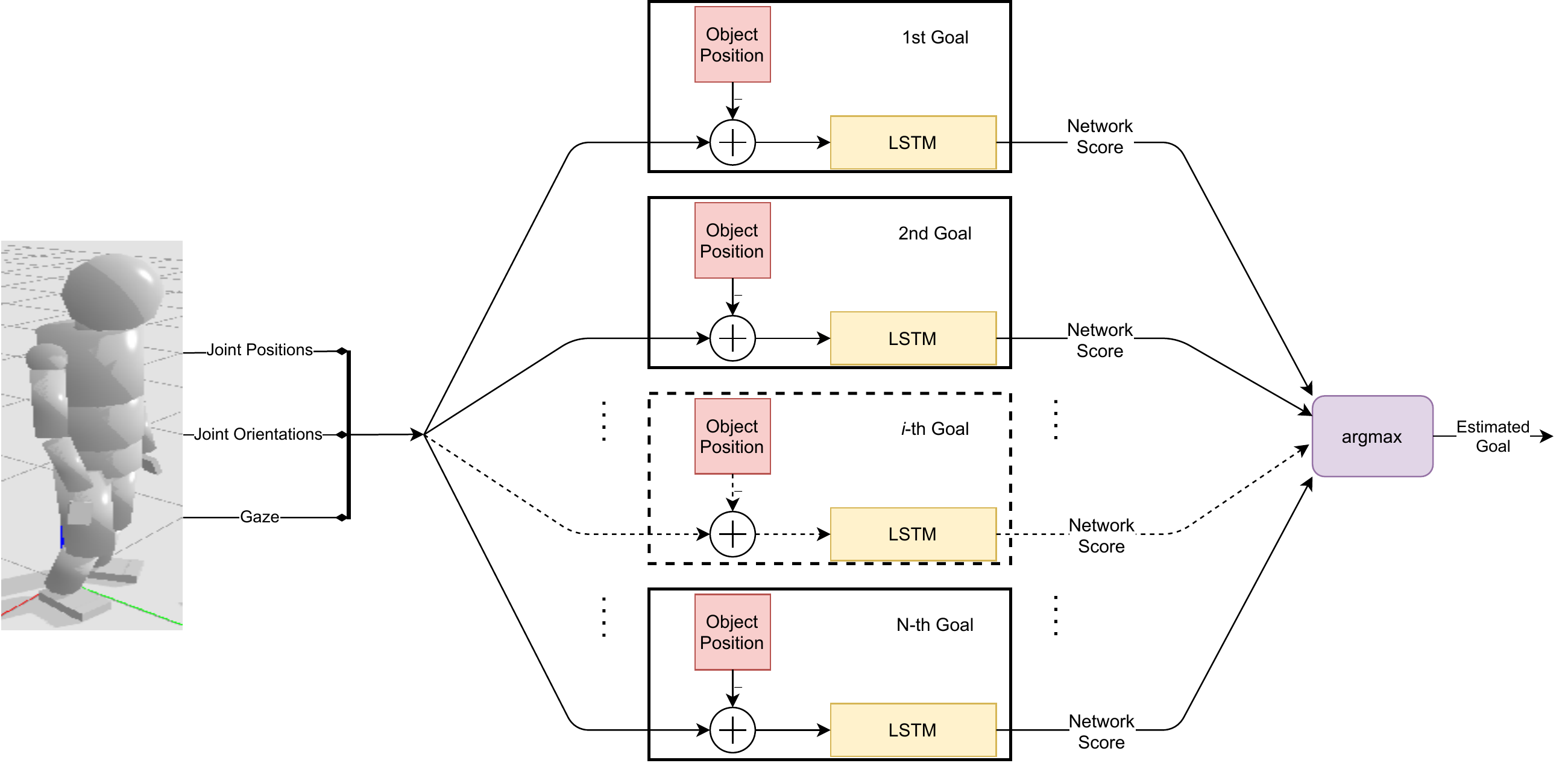}
\caption{Proposed human action prediction pipeline. We evaluate observed sequences with respect to each goal using an ensemble of identical LSTM networks as classifiers. The network with the largest score is considered to be associated with the current prediction of the subject's goal.}
\label{fig:architecture}
\vspace{-0.7cm}
\end{figure}

Given that, we have decided to train only a single LSTM model to classify whether the input sequence of features is the pertaining goal or not.
This approach enables us to easily add or remove goals and could potentially lead to easier general application.
The selected model has an input layer, an LSTM layer with 20 hidden units, a fully connected layer, a softmax layer, and a classification layer.
Every 10 frames it takes the whole history of selected input features labeled as \{0, 1\} and outputs a score between [0, 1] -- network's confidence in the estimated action.
Joint positions are used to calculate Euclidean distances towards each of the $N$ goals in the dataset, and gaze and orientation vectors are used to calculate the distance between them and the vector pointing towards the position of each goal.
All of the features are normalized based on the average value on the training set.
During runtime, we evaluate all of the selected features for each of the $N$ goals and send them as inputs to the ensemble of $N$ identical LSTM networks, where $N$ is the number of objects -- for the MoGaze dataset $N = 10$.
We compare the output of each of the networks and select the goal whose network has the maximum score.
An illustration of this process is shown in Fig.~\ref{fig:architecture}.

The main contributions of the current work lie in feature selection for human action prediction and demonstration that the problem at hand can be successfully approached using LSTMs based on selected features.
Because of that, hyperparameters and the layout of the proposed LSTM network have not been further scrutinized.
\subsection{Feature Selection}
\label{sec:features}
\begin{figure}[t!]
\centering
\includegraphics[clip, trim=0cm 4.2cm 1cm 4.2cm,width =\textwidth]{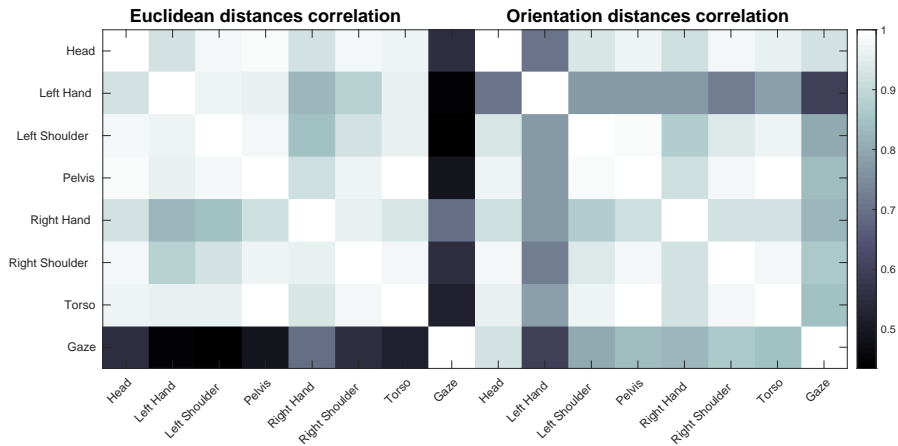}
\caption{Correlations of various dataset features. Both orientations and Euclidean distances of hands correlate poorly with the other joints distances.}  \vspace{-0.5cm}
\label{fig:correlation}
\end{figure}
The LSTM networks excel at capturing long-term dependencies of input signals with the cost of adding a substantial amount of parameters to the model.
Because of that, we propose a method for feature selection to reduce the input size and enable a quicker performance of the model.
We will demonstrate proposed feature selection on the MoGaze dataset which has recorded positions and orientations of multiple joints.
Our analysis will be focused on the following joints: head, torso, pelvis, both hands and both shoulders.
We selected these joints based on the intuition that other joints, such as knees and elbows, do not contain any additional information for inferring which object the human is about to pick.
Following the preliminary results published in \cite{kratzer2020mogaze}, wherein the authors asserted that gaze is the most powerful predictor of human action, we compared correlations of our selected signals and the gaze, and the results can be seen in Fig.~\ref{fig:correlation}.

In the MoGaze dataset participants were advised to use only one hand for performing tasks \cite{kratzer2020mogaze}, resulting in heavily favored use of the right hand.
As can be seen in Fig.~\ref{fig:correlation}, the position and orientation distances of the left hand correlate poorly with other joints; however, pelvis, shoulders, head and torso orientation distances correlate reasonably high with each other.
Such high correlations suggest that only one of the features could be selected as an input to the network.
Furthermore, we also calculated the cumulative gaze feature which at frame $t$ takes the value of the sum of the number of frames that the gaze towards the goal was under the threshold of $0.1$\,m.
Our motivation in constructing this feature lies in the intuition that during the action, human fixes gaze for a prolonged period of time on the goal it wants to visit, and it would be useful to record the entire history of that signal.
In order to determine which of the aforementioned cues could act as a good action predictor, we compared the individual accuracy of all of the selected features.
These single cue predictor results are called \emph{baselines}, since any other more complex model should obtain better results.
We followed the accuracy evaluation for a time span of 3 seconds, with the grasp of the object occurring at the third second, as proposed in \cite{kratzer2020mogaze}.
\begin{figure}[t]
\centering
\includegraphics[clip, trim=0.5cm 3.5cm 0.5cm 4.cm, width =1 \textwidth  ]{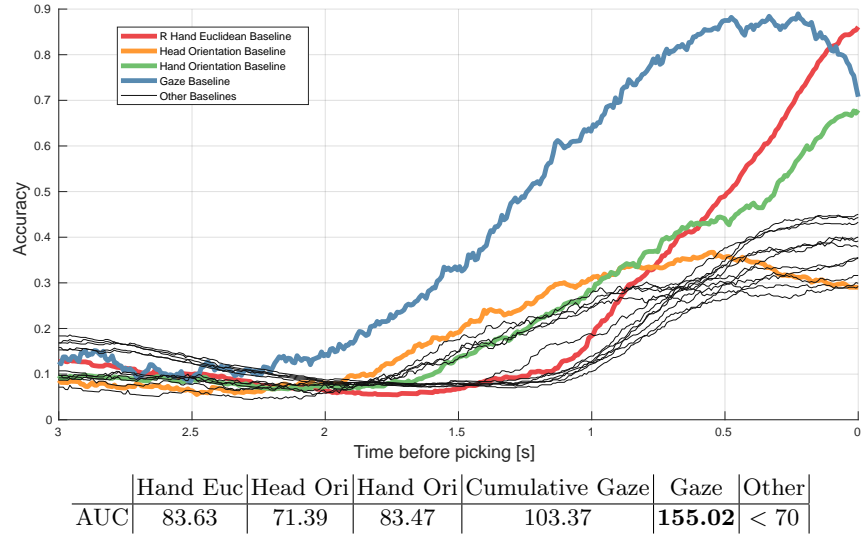}
\begin{tabular}{ c | c | c |c  |c| c |c |c| c}
 & Hand Euc & Head Ori & Hand Ori & Cumulative Gaze & Gaze & Other\\ \hline
AUC & 83.63 & 71.39 & 83.47 & 103.37 & \textbf{155.02} & $<70$  \\
\end{tabular}
\centering
  \caption{Comparison of AUCs for all baselines.
Larger AUC value indicates that the baseline is a better predictor of action.
The gaze baseline clearly outperforms other baselines until the last few moments when the proximity of the reaching hand becomes a better indicator of the picked object.}
\label{fig:baselines}
\vspace{-0.5cm}
\end{figure}

Fig.~\ref{fig:baselines} presents the calculated area under the accuracy curve (AUC) for all of the baselines.
We decided to use the AUC score evaluation from the computer vision applications such as face alignment \cite{trigeorgis2016mnemonic}, where the authors calculate the area under the accumulated error distribution curve,  which resembles our accuracy curves.
We find the AUC score as a reasonable scalar value containing information about the accuracy during the whole prediction period.
In the MoGaze dataset, information is sampled at 120 Hz, which means that maximum AUC for the three second sequence is 360, while the expected AUC of a random goal guesser for 10 goals would be 36.
This gives us border values for accuracy evaluation.
Based on the AUC scores shown in Fig.~\ref{fig:baselines}, we selected the following reduced feature set: gaze, cumulative gaze, head orientation, hand position and hand orientation.

\section{Experimental Results}
\label{sec:results}
In this section we describe in detail the MoGaze dataset that we used for training and evaluation.
Furthermore, we discuss the results and present several versions of the LSTM models that we designed for human action prediction.

\subsection{Dataset and Training}
The MoGaze dataset includes a total of 180 minutes of motion capture data with 1627 pick and place actions \cite{kratzer2020mogaze}.
For one participant (participant no. 3) the eye-tracker device did not work so we excluded this session leaving a total of 1435 picking segments.
Each group of segments is preceded by the instructions to the participant, e.g. \textit{"Set the table for 2 persons"}, but we do not use this information.
Each segment consists of multiple frames before the actual picking happens and a labeled of the object that is eventually picked.
Frames in which an object is being carried are discarded.
It might be beneficial to additionally discard the parts of segments before the instruction has been given as well as those involving moving of chairs; however, we decided to leave the dataset intact for easier future comparison.
In order to prepare the dataset for model training, we split the data into training and testing datasets.
The training dataset consisted of sessions with subjects 1-2 and 4-5 (a total of 853 segments), while the testing dataset contained sessions with subjects 6-7 (a total of 582 segments).
We have performed feature selection as in Section~\ref{sec:features} and trained the network using MATLAB with Adam optimization  \cite{kingma2014adam} training with 5 epochs and batch size of 5.
\subsection{Experimental Results}
\begin{figure}[t!]
\centering
\begin{subfigure}{.45\textwidth}
	\includegraphics[clip, trim=3.8cm 7.8cm 4.2cm 7.6cm, width =\textwidth  ]{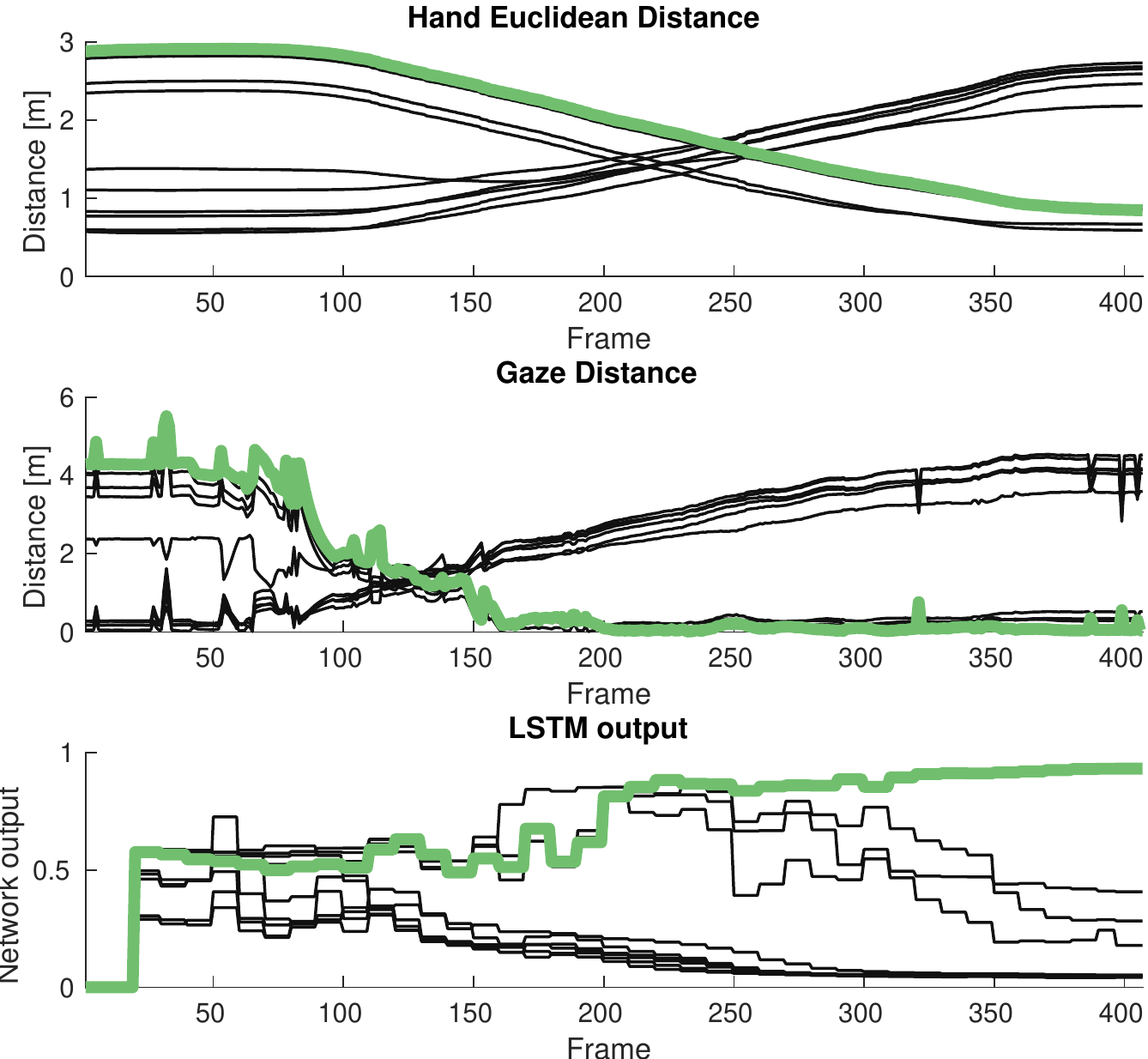}
\caption{}
\end{subfigure}
\begin{subfigure}{.45\textwidth}
\includegraphics[clip, trim=3.5cm 7.4cm 4.2cm 7.3cm, width = \textwidth  ]{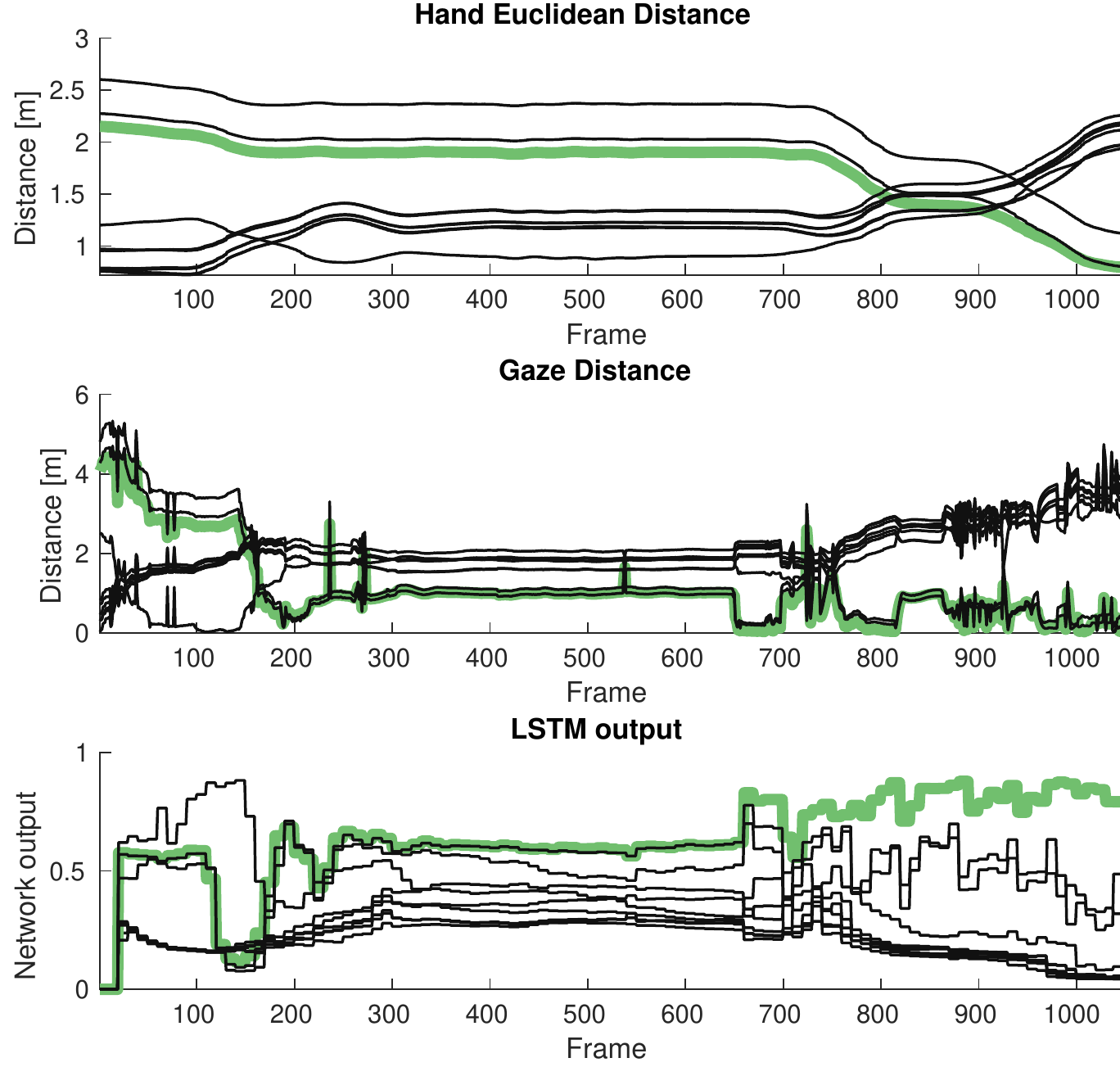}
\caption{}
\end{subfigure}\\ \vspace{2mm} \hspace{-2mm}
\begin{subfigure}{.45\textwidth}
\includegraphics[clip, trim=3.8cm 7.8cm 4.2cm 7.6cm, width = \textwidth  ]{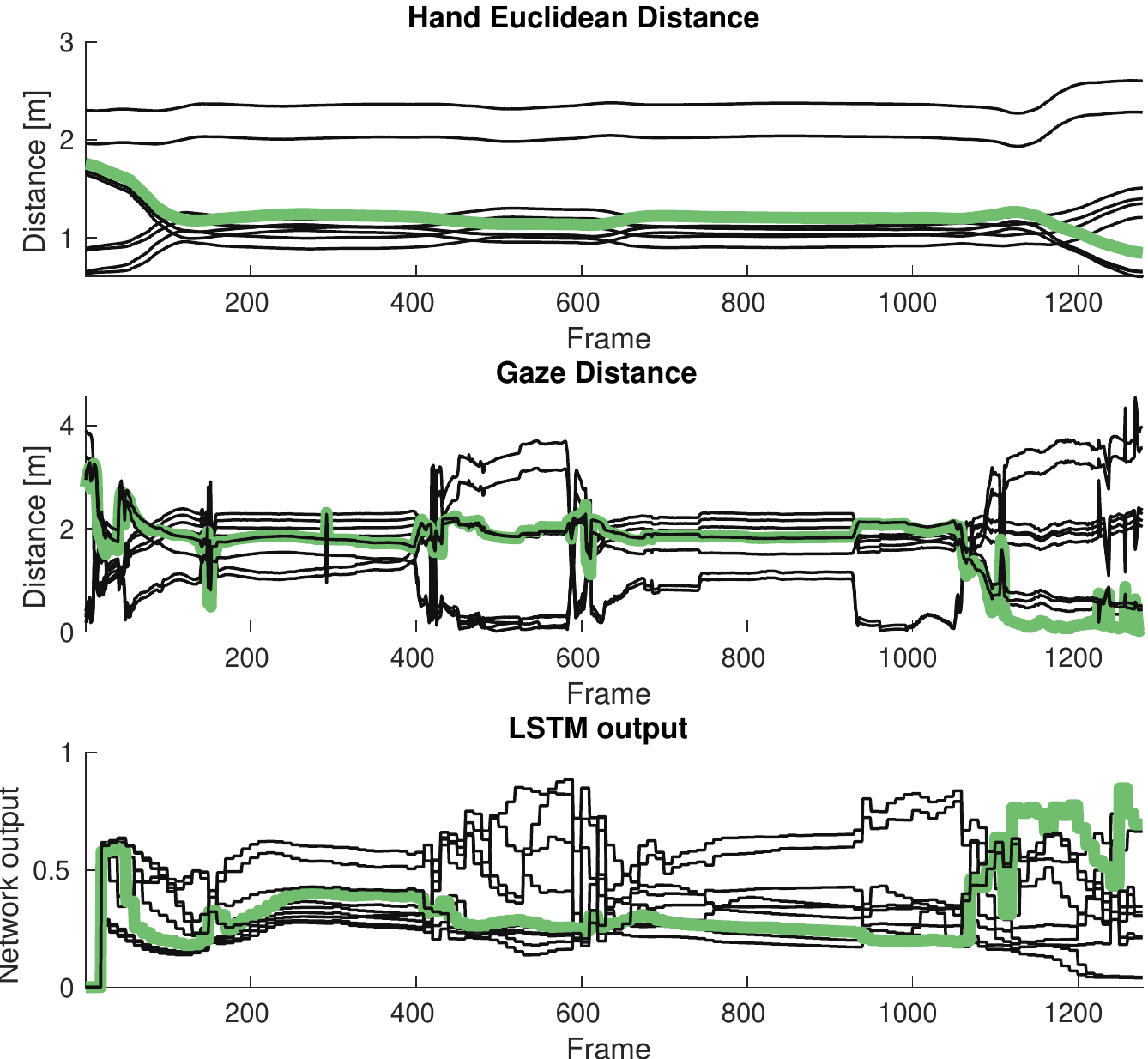}
\caption{}
\end{subfigure}
\begin{subfigure}{.45\textwidth}
\includegraphics[clip, trim=3.8cm 7.8cm 4.2cm 7.6cm, width = \textwidth  ]{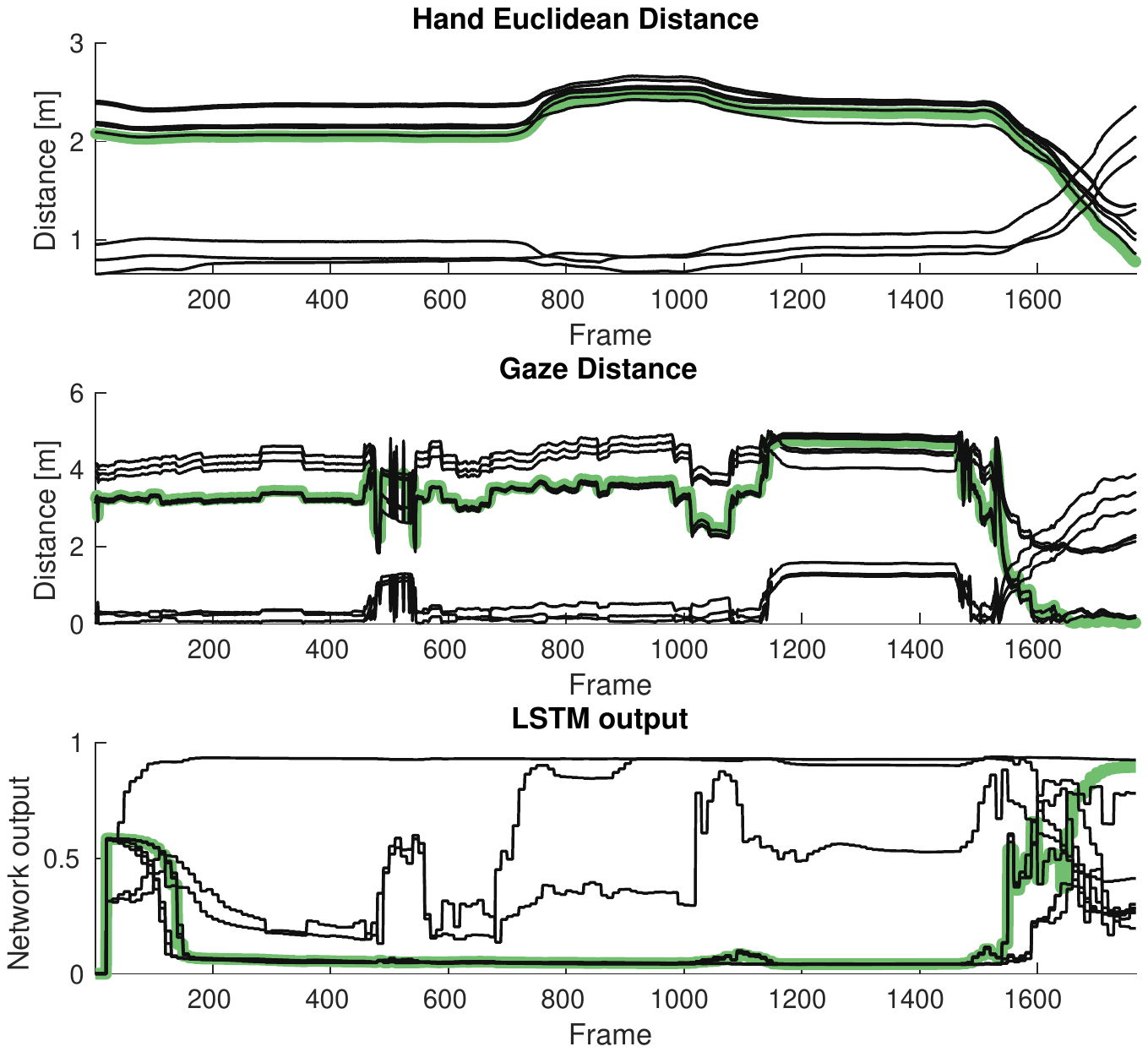}
  \caption{}
\end{subfigure}\\
  \caption{Examples of selected model inputs and outputs. The feature corresponding with the object being picked is colored green while the black lines denote features associated with other objects.}
\label{fig:examples}
\vspace{-0.7cm}
\end{figure}

For accuracy evaluation, we used AUC for the last three seconds of each experiment.
Since we want our network to be applicable in real time applications, all the inputs are recorded each 10 steps (12Hz) and network outputs are considered to be valid until the next iteration.
\begin{figure}[t]
\centering
\includegraphics[clip, trim=0.5cm 3.5cm 0.5cm 4.5cm, width =1 \textwidth  ]{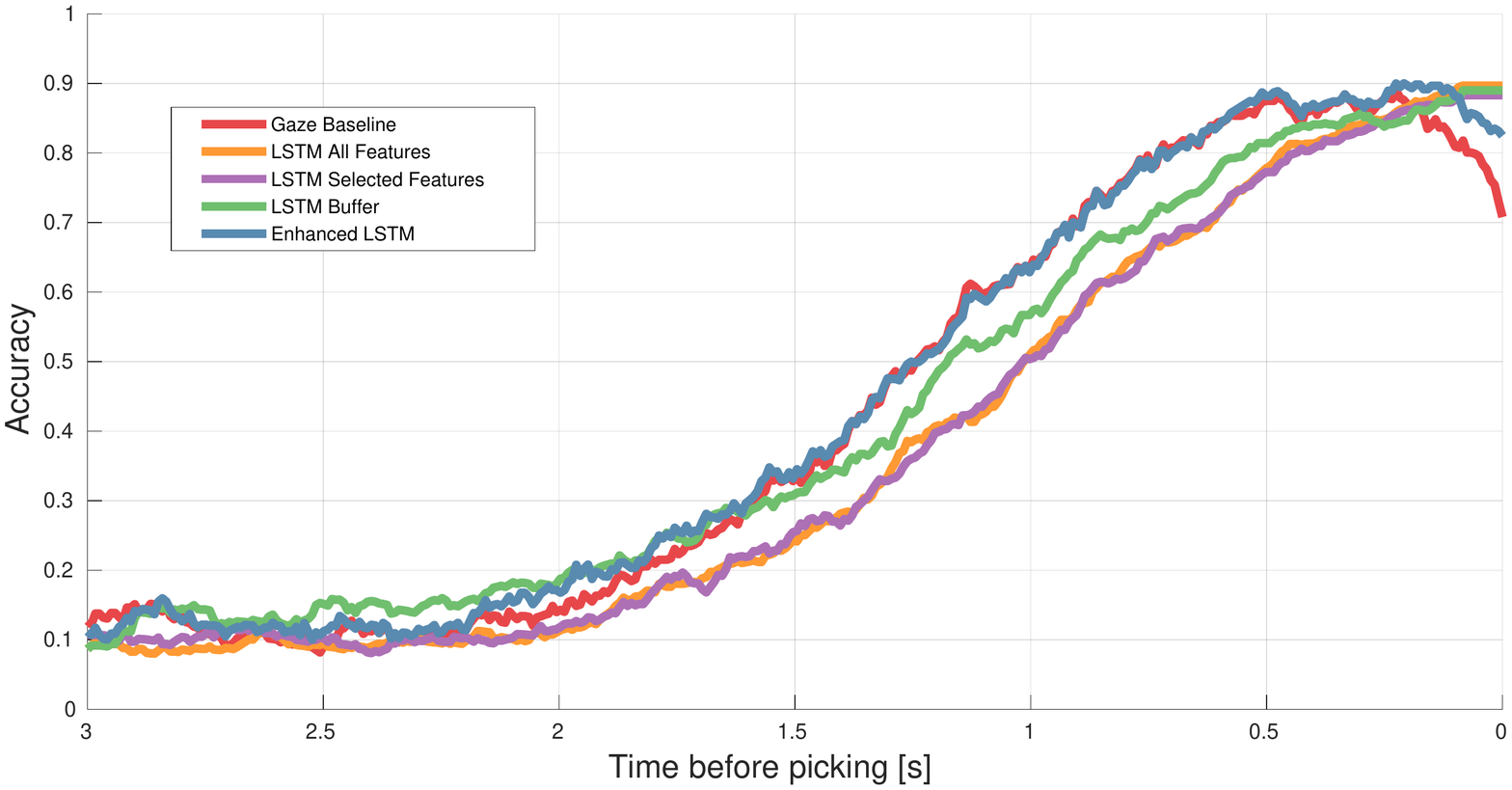}
\begin{tabular}{ c | c | c |c  |c| c |c |c| c}
 & Gaze & LSTM & LSTM Select  &LSTM Buff & Enhanced LSTM\\ \hline
AUC & 155.02 & 134.24 & 134.43 & 151.98 &\textbf{159.01}  \\
\end{tabular}
\centering
  \caption{Proposed models' performances. Eye-gaze has proven to be strongest indicator of human action on this dataset.
}
\vspace{-0.5cm}
\label{fig:results}
\end{figure}
First, we trained and tested the proposed network using the whole signal history of all recorded joints' orientations and distances as inputs, including the gaze (we dubbed this version simply as \emph{LSTM}).
With the AUC of 134.24, the LSTM model performed worse than the gaze baseline (155.02) during majority of experiments and only succeeded to beat it in the last few moments.
On top of that, average prediction time was too slow to work with the aimed frequency of 12 Hz.
Therefore, we further tried to improve the results and execution time by reducing the input feature set as discussed in Section~\ref{sec:features} (dubbed \emph{LSTM Select}).
While it slightly improved execution time, the AUC remained unchanged for the LSTM Select model.
In order to reduce the run time, we decided to reduce the complexity by using a buffer to send only the last 20 frames to the network (dubbed \emph{LSTM Buff}).
The LSTM Buff model had a satisfying run-time and achieved AUC of 151.98; however, its score was still lower than that of the gaze baseline.
As tuning the network hyperparameters mentioned at the beginning of this section did not result in any considerable improvements, we decided to further analyze the inputs and the outputs of the proposed model to see where accuracy could be increased and the results are shown in Fig.~\ref{fig:examples}.
For example, in Fig.~\ref{fig:examples}a), one can see that our model has clearly distinguished which object is the goal after step 250, while Euclidean distance is not the lowest at any point.
Also, the gaze signal corresponding with the picked goal has a spike around frame 325.
This did not have major effect on the output of the proposed model.
In Fig.~\ref{fig:examples}b), the proposed model predicted high probability of the picked object very early but only succeeded to isolate it from nearby objects after step 700.
Gaze and Euclidean distance alone could not make such distinction.
In contrast, in Fig.~\ref{fig:examples}c) and Fig.~\ref{fig:examples}d), proposed model behaved poorly.
One possible explanation could be that the subject looked at other objects without moving towards any of the goals, thus not giving the model enough information to conclude the exact goal until the end of the segment.

Given the previous analysis, we can see that the gaze baseline acts as the strongest predictor of the object that the human is going to pick, and furthermore, it can distinguish the actual goal among the nearby objects with pinpoint accuracy -- something that our network model struggles with.
The fact that the gaze is such a strong indicator of action is not surprising; indeed, visual fixation is necessary for object identification and comes after the brain automatically and in parallel gathers basic features, such as colors, shape, and motion \cite{Treisman1980}.
Given that, we have enhanced our model with a few simple conditions.
If the highest LSTM score is larger than the \textit{score threshold}, we consider that object to be the goal.
Otherwise, we check if the subject is looking directly at any of the objects, by comparing the minimum gaze distance to the objects with the \textit{gaze threshold}.
If the subject is looking at the object, we output that goal, otherwise we select the highest LSTM score.
This enhancement led to 159.01 AUC, slightly improving over the gaze baseline by $2.6\%$.
We used the score threshold of 0.49 and gaze threshold of 0.2, which have been selected using grid search on the training set with the AUC as the target function.
At the time of writing this paper, the only method for human action prediction on the MoGaze dataset was published in \cite{kratzer2020mogaze}.
The method is based on RNNs but the authors do not provide implementation details rendering a direct comparison difficult.
However, by qualitatively comparing the accuracy figures, we can assert that the proposed approach seems to yield more accurate action prediction results.

\begin{figure}[t!]
\centering
\begin{subfigure}{0.45\textwidth}
\includegraphics[clip,  trim=3.5cm 7.5cm 4cm 8cm, width = \textwidth  ]{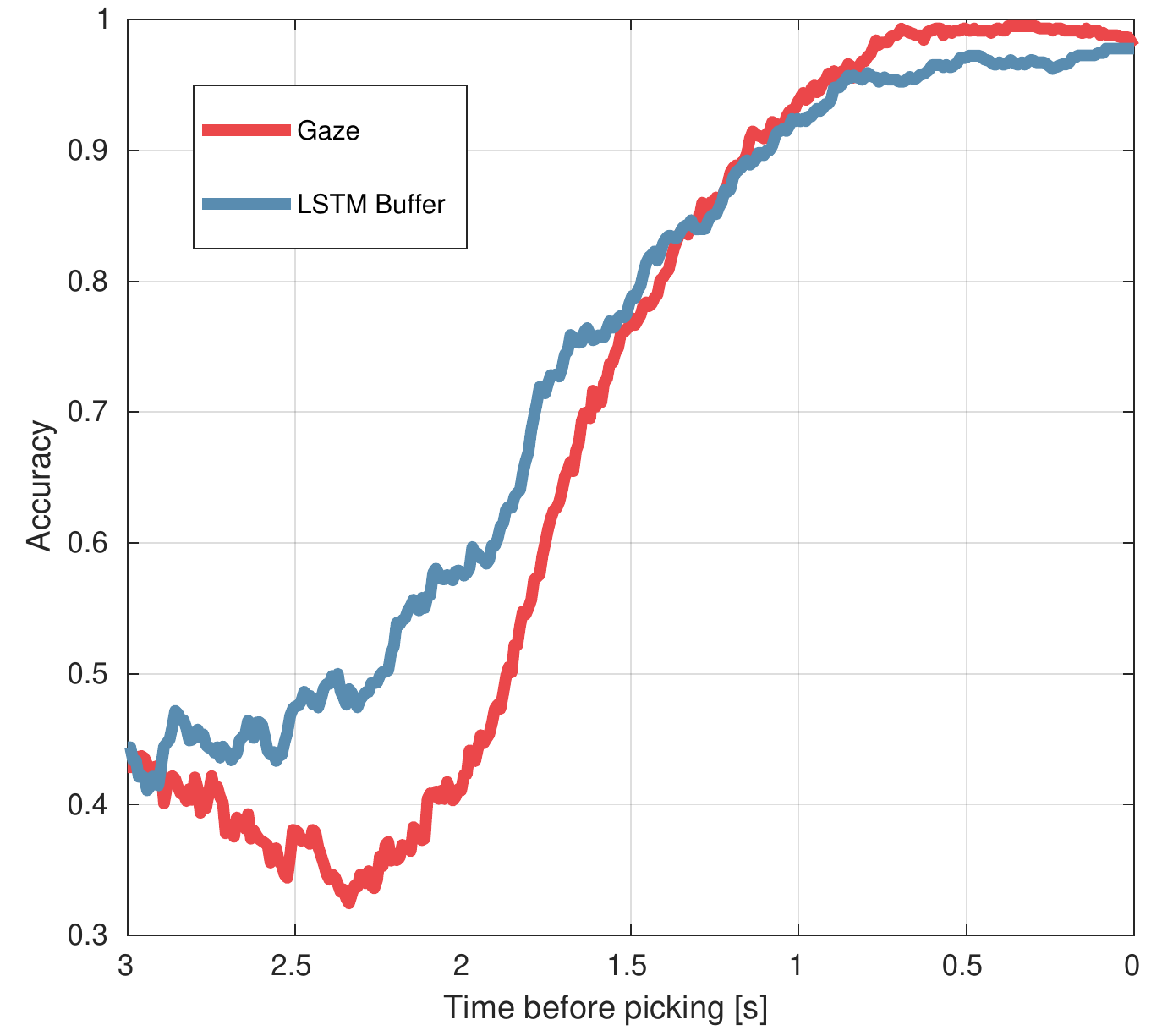}
  \caption{Network performance distinguishing between specific locations.} 
\end{subfigure}
\begin{subfigure}{0.45\textwidth}
  \centering
\includegraphics[clip, trim=3.5cm 7.5cm 4cm 8cm, width = \textwidth  ]{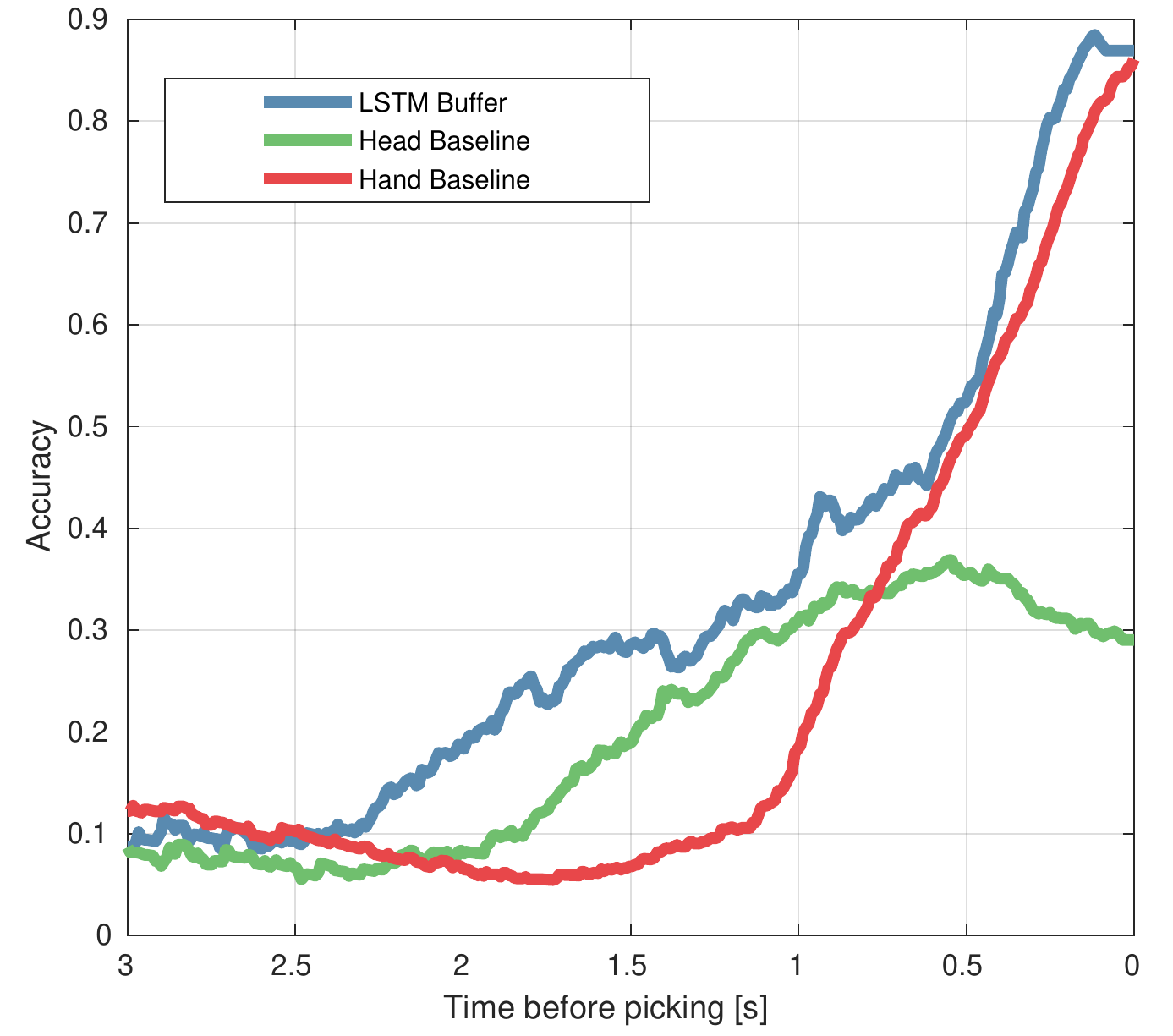}
  \caption{Network performance without eye-gaze.}
\end{subfigure}
  \caption{The proposed action prediction model has better performance than any baseline if the gaze feature is not available or when one only needs to determine the macro location of the object being picked.}
\label{fig:result}
\vspace{-0.7cm}
\end{figure}

Although our LSTM models underperformed or showed only a slightly better result than the gaze baseline, we have observed that the LSTMs are better at identifying the macro objects from which the human is going to pick a specific object.
Namely, all objects in the Mogaze dataset are placed on three specific macro locations: the table or two shelves.
By analyzing the estimated macro location accuracy, the LSTM Buff model scored 266.31 AUC, while the gaze baseline scored 251.31 AUC, as shown in Fig.~\ref{fig:result}a).
A possible explanation could be that test subjects had to naturally look around while moving towards the goal, preventing them from keeping the gaze fixed on the correct macro location.
Our model, on the contrary, managed to successfully leverage the motion cues towards the goal and obtained a higher score in identifying the macro locations.

Since gaze is the most powerful predictor of human action, it begs the question of what kind of performance could be achieved if such a cue was not available?
Such a question could be further motivated since presently gaze tracking is done using a cumbersome apparatus that can impede subject's efficiency and comfort in real-life applications.
Thus it would be interesting to test what performances could be achieved if the gaze information is not available.
We tested the behavior of the proposed LSTM Buff giving only hand and head positions and orientations as input.
Because of the reduced number of features, we have increased the number of hidden units to 40 for this model.
Our model successfully combined those signals and scored AUC of 118.60 which is considerably higher than the baselines (83.63 and 71.39), as can be seen in Fig.~\ref{fig:result}b).
\section{Conclusion and Future work}
\label{sec:conclusion}
In this paper, we have proposed a novel human action prediction model based on the ensemble of LSTM networks.
We have thoroughly analyzed the dataset and selected appropriate cues for human action prediction.
The identical copies of LSTM networks receive distances and orientations of selected human joints towards each of the 10 goals as recorded.
The proposed framework is trained to classify whether the observed sequence belongs to an object that the human is going to pick or not.
During the runtime, we compare the network outputs for each object and select the one with the highest score as our prediction of the human action.
We have exhaustively evaluated the proposed method and the experiments demonstrated its effectiveness in comparison to single cue baselines.

For future work we plan to analyze possible improvements and alternative architectures of the proposed network as well as demonstrate its generalization power by testing it on our own dataset.
\section*{Acknowledgment}
This research has been supported by the European Regional Development Fund under the grant KK.01.1.1.01.0009 (DATACROSS).
%

\bibliographystyle{abbrv}
\bibliography{library}

\end{document}